\def\year{2019}\relax
\documentclass[letterpaper]{article}
\usepackage{aaai19}
\usepackage{times}
\usepackage{amsmath,amssymb,amsfonts}
\usepackage{algorithmic}
\usepackage{graphicx}
\usepackage{textcomp}
\usepackage{xcolor}
\usepackage{booktabs} 
\usepackage{xcolor}
\usepackage{soul}
\usepackage{bm}
\usepackage{helvet}  
\usepackage{courier}  
\usepackage{url}  
\usepackage{graphicx}  
\usepackage{subcaption}
\usepackage{courier}
\usepackage{multirow}
\usepackage{color}
\usepackage{todonotes}
\usepackage{amssymb}
\usepackage{enumitem}
\usepackage{multirow}
\usepackage[linesnumbered, ruled]{algorithm2e}
\usepackage{amsmath}

\newcommand{\nop}[1]{}

\begin{document}
\title{Revisiting Spatial-Temporal Similarity: A Deep Learning Framework\\ for Traffic Prediction}
\author{Huaxiu Yao\thanks{Equal contribution}, Xianfeng Tang\footnotemark[1], Hua Wei, Guanjie Zheng, Zhenhui Li\\Pennsylvania State University\\\{huaxiuyao, xianfeng, hzw77, gjz5038, jessieli\}@ist.psu.edu}
\maketitle
\begin{abstract}
	Traffic prediction has drawn increasing attention in AI research field due to the increasing availability of large-scale traffic data and its importance in the real world. For example, an accurate taxi demand prediction can assist taxi companies in pre-allocating taxis. The key challenge of traffic prediction lies in how to model the complex spatial dependencies and temporal dynamics. 
	Although both factors have been considered in modeling, existing works make strong assumptions about spatial dependence and temporal dynamics, i.e., spatial dependence is stationary in time, and temporal dynamics is strictly periodical. However, in practice the spatial dependence could be dynamic (i.e., changing from time to time), and the temporal dynamics could have some perturbation from one period to another period. 
	In this paper, we make two important observations: (1) the spatial dependencies between locations are dynamic; and (2) the temporal dependency follows daily and weekly pattern but it is not strictly periodic for its dynamic temporal shifting. To address these two issues, we propose a novel Spatial-Temporal Dynamic Network (STDN), in which a flow gating mechanism is introduced to learn the dynamic similarity between locations, and a periodically shifted attention mechanism is designed to handle long-term periodic temporal shifting. To the best of our knowledge, this is the first work that tackle both issues in  a unified framework. Our experimental results on real-world traffic datasets verify the effectiveness of the proposed method. 
\end{abstract}

\section{Introduction}
Traffic prediction - a spatial temporal prediction problem, has drawn increasing attention due to the growing traffic related datasets and for its impacts in real-world applications.. In the meantime, an accurate traffic prediction model is essential to many real-world applications. For example, taxi demand prediction can help taxi companies pre-allocate taxis; traffic volume prediction can help transportation department better manage and control the traffic to ease traffic congestion.

In a typical traffic prediction setting, given historical traffic data (e.g., traffic volume of a region or a road intersection for each hour during the previous month), one needs to predict the traffic for the next time slot. A number of studies have investigated traffic prediction for decades. In time series community, autoregressive integrated moving average (ARIMA) and Kalman filtering have been widely applied to traffic prediction problems~\cite{li2012prediction,moreira2013predicting,shekhar2008adaptive,lippi2013short}. While these earlier methods study traffic time series for each individual location,  separately, recent studies started taking into account spatial information (e.g., adding regularizations on model similarity for nearby locations)~\cite{deng2016latent,ide2011trajectory,zheng2013time} and external context information (e.g., adding features of venue information, weather condition, and local events)~\cite{wu2016interpreting,pan2012utilizing,tong2017sim}. However, these approaches are still based on traditional time series models or machine learning models and do not well capture the complex non-linear spatial-temporal dependency. 

Recently, deep learning has made achieved tremendous success in many challenging learning tasks~\cite{lecun2015deep}. 
The success has inspired several studies to apply deep learning techniques to traffic prediction problem. For example, several studies~\cite{zhang2016deep,zhang2016dnn,ma2017learning} have modeled city-wide traffic as a heatmap image and use convolutional neural network (CNN) to model the non-linear spatial dependency. To model non-linear temporal dependency, researchers propose to use recurrent neural network (RNN)-based framework~\cite{yu2017deep,cuideep}. \citeauthor{yao2018deep} further propose a method to jointly model both spatial and temporal dependencies by integrating CNN and long short-term memory (LSTM)~\cite{yao2018deep}.

Although both spatial dependency and  temporal dynamics have been considered in deep learning for traffic prediction, the existing method have two major limitations. First, the spatial dependency between locations relies only on the similarity of historical traffic~\cite{zhang2016deep,yao2018deep} and the model learns a static spatial dependency. However, the dependencies between locations could change over time. For example, in the morning, the dependency between a residential area and a business center could be strong; whereas in late evening, the relation between these two places might be very weak. However, such dynamic dependencies have not been considered in previous studies. 

Another limitation is that many existing studies ignore the shifting of long-term periodic dependency. Traffic data show a strong daily and weekly periodicity and the dependency based on such periodicity can be useful for prediction. However, one challenge is that the traffic data are not strictly periodic. For example, the peak hours on weekdays  usually happen in the late afternoon, but could vary from 4:30pm to 6:00pm on different days. Though previous studies~\cite{zhang2016deep,zhang2016dnn} consider periodicity, they fail to consider the sequential dependency and the temporal shifting in the periodicity. 

To address the aforementioned challenges, we propose a novel deep learning architecture, \textbf{S}patial-\textbf{T}emporal \textbf{D}ynamic \textbf{N}etwork (\textbf{STDN}) for traffic prediction. STDN is based on a spatial-temporal neural network, which handles spatial and temporal information via local CNN and LSTM, respectively. A flow-gated local CNN is proposed to handle spatial dependency by modeling the dynamic similarity among locations using traffic flow information. A periodically shifted attention mechanism is proposed to learn the long-term periodic dependency. The proposed mechanism captures both long-term periodic information and temporal shifting in traffic sequence via attention mechanism. Furthermore, our method uses LSTM to handle the sequential dependency in a hierarchical way. 

We evaluate the proposed method on large-scale real-world public datasets including taxi data of New York City (NYC) and bike-sharing data of NYC. The comprehensive comparisons with the state-of-the-art methods demonstrate the effectiveness of our proposed method. Our contributions are summarized below: 
\begin{itemize}[leftmargin=*]
	\item We propose a flow gating mechanism to explicitly model dynamic spatial similarity. The gate controls information propagation among nearby locations.
	\item We propose a periodically shifted attention mechanism by taking long-term periodic information and temporal shifting simultaneously.
	\item We conduct experiments on several real-world traffic datasets. The results show that our model is consistently better than other state-of-the-art methods.
\end{itemize}

\section{Related Work}
Data-driven traffic prediction problems have received wide attention for decades. Essentially, the aim of traffic prediction is to predict a traffic-related value for a location at a timestamp based on historical data. In this section, we discuss the related work on traffic prediction problems.

In time series community, autoregressive integrated moving average (ARIMA), Kalman filtering and their variants have been widely used in traffic prediction problem~\cite{shekhar2008adaptive,li2012prediction,moreira2013predicting,lippi2013short}. Recent studies further explore the utilities of external context data, such as venue types, weather conditions, and event information~\cite{pan2012utilizing,wu2016interpreting,rong2017taxi}. In addition, spatial information has also been explicitly modeled in recent studies~\cite{deng2016latent,tong2017sim,ide2011trajectory,zheng2013time}. \emph{However, all of these methods fail to model the complex nonlinear relations of the space and time.}

Deep learning models provide a new promising way to capture non-linear spatiotemporal relations, which have achieved great success in computer vision and natural language processing~\cite{lecun2015deep}. In traffic prediction, a series of studies have been proposed based on deep learning techniques. The first line of studies stacked several fully connected layers to fuse context data from multiple sources for predicting traffic demand ~\cite{wei2016zest}, taxi supply-demand gap~\cite{wang2017deepsd}. \emph{These methods used extensive features, but do not model the spatial and temporal interactions explicitly.} 

The second line of studies applied convolutional structure to capture spatial correlation for traffic flow prediction~\cite{zhang2016deep,zhang2016dnn}. The third line of studies used recurrent-neural-network-based model for modeling sequential dependency~\cite{yu2017deep,cuideep}. \emph{However, while these studies explicitly model temporal sequential dependency or spatial dependency, none of them consider both aspects simultaneously.}

Recently, several studies use convolutional LSTM~\cite{xingjian2015convolutional} to handle spatial and temporal dependency for taxi demand prediction~\cite{ke2017short,zhou2018predicting}. Yao et al. further proposed a multi-view spatial-temporal network for demand prediction, which learns the spatial-temporal dependency simultaneously by integrating LSTM, local-CNN and semantic network embedding~\cite{yao2018deep}. Based on road network, several studies extended traditional CNN and RNN structure to graph-based CNN and RNN for traffic prediction, such as graph convolutional GRU~\cite{li2017graph}, graph attention~\cite{zhang2018gaan}.~\emph{In these studies, the similarity between regions is based on static distance or road structure. They also overlook the long-term periodic influence and temporal shifting in time series prediction.} 

In summary, our proposed model explicitly handle dynamic spatial similarity and temporal periodic similarity jointly via flow gating mechanism and periodically shifted attention mechanism, respectively. 

\section{Notations and Problem Formulation}
\label{sec:problem}
We split the whole city to an $a\times b$ grid map with $n$ regions in total ($n = a \times b$), and use $\{1,2,\dots, n\}$ to denote them. We split the whole time period (e.g., one month) into $m$ equal-length continuous time intervals. The moving trip of any individual, which is an essential part of the entire citywide traffic, always departs from a region, and arrives at the destination one after a while. We define the start/end traffic volume for a region as the number of trips departing/arriving from/in the region during a fixed time interval. Formally, $y_{i,t}^s$ and $y_{i,t}^e$ stand for the start/end traffic volume for region $i$ during the $t$-th time interval. Moreover, aggregation of individual trips formulates the traffic flow, which describes time-enhanced movements between certain pair of regions. Formally, the traffic flow starting from region $i$ in time interval $t$ and ending in region $j$ in time interval $\tau$ is denoted as $f_{i,t}^{j,\tau}$. Obviously, the traffic flow reflects region-wise connectivity, as well as the propagation of individuals. The illustration of traffic volume and flow are given in Figure \ref{fig:framework}(c).

\textbf{Problem (Traffic Volume Prediction)} Given the data until time interval $t$, the traffic volume prediction problem aims to predict the start and end traffic volume at time interval $t+1$.  

\section{Spatial-Temporal Dynamic Network}
\label{sec:model}
\begin{figure*}[!t]
	\centering
	\includegraphics[width=0.75\textwidth]{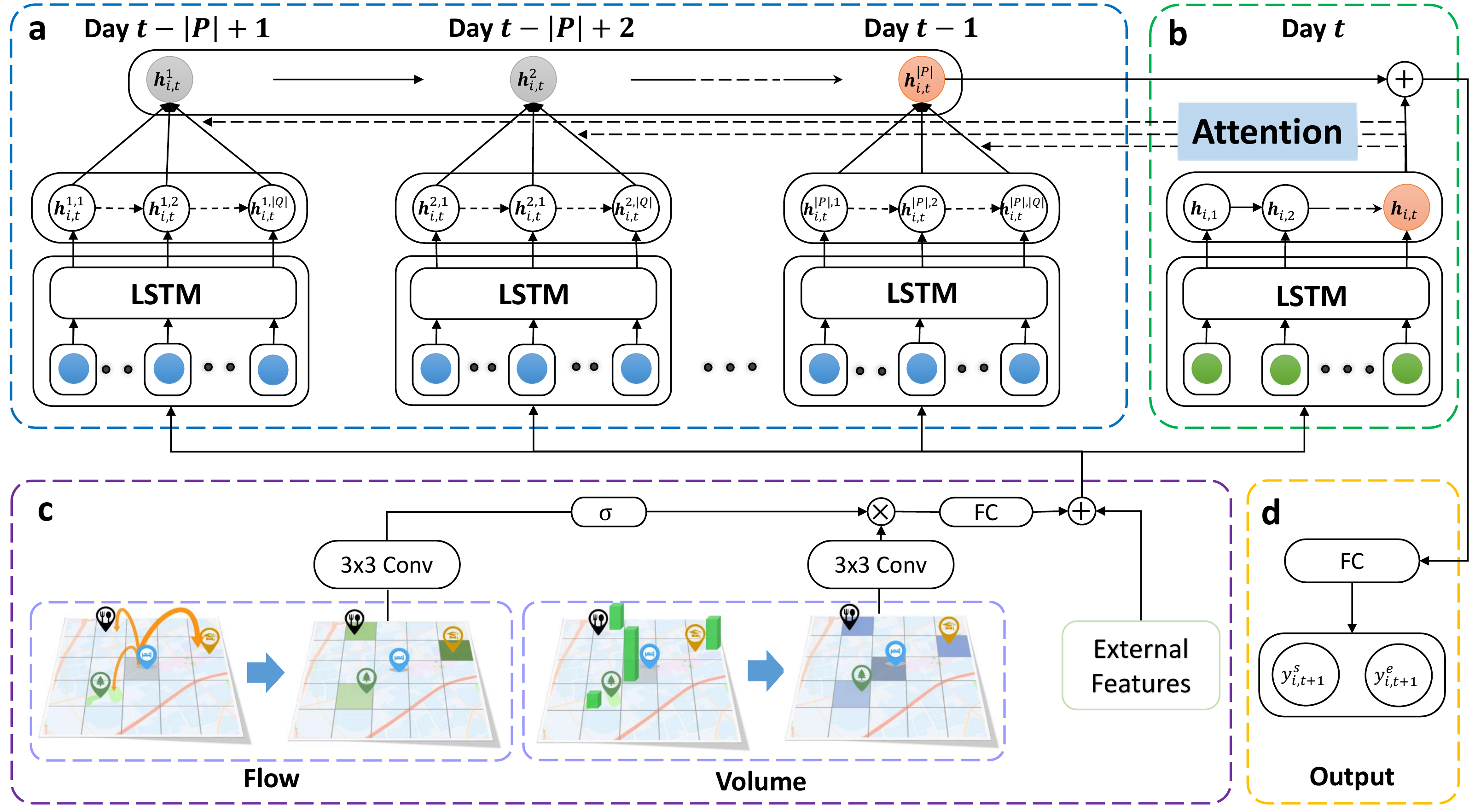}
	\caption{The architecture of STDN. 
		(a) Periodically shifted attention mechanism captures the long-term periodic dependency and temporal shifting. For each day, we also use LSTM to capture the sequential information. (b) The short-term temporal dependency is captured by one LSTM. (c) The flow gating mechanism tracks the dynamic spatial similarity representation by controlling the spatial information propagation; FC means fully connected layers and Conv means several convolutional layers. (d) A unified multi-task prediction component predicts two types of traffic volumes simultaneously.}
	\label{fig:framework}
\vspace{-1em}
\end{figure*}

In this section, we describe the details for our proposed \textbf{S}patial-\textbf{T}emporal \textbf{D}ynamic \textbf{N}etwork (\textbf{STDN}). 
Figure~\ref{fig:framework} shows the architecture of our proposed method. 

\subsection{Local Spatial-Temporal Network}
In order to capture spatial and temporal sequential dependency, combining local CNN and LSTM has shown the state-of-the-art performance in taxi demand prediction~\cite{yao2018deep}. Here, we also use local CNN and LSTM to deal with spatial and short-term temporal dependency. In order to mutually reinforce the prediction of two types of traffic volumes (i.e., start and end volumes), we integrate and predict them together. This part of our proposed model is called Local Spatial-Temporal Network (LSTN).
\label{sec:localst}
\subsubsection{Local spatial dependency}
\label{sec:localspatial}
Convolutional neural network (CNN) is used to capture the spatial interactions. Suggested in~\cite{yao2018deep}, treating the entire city as an image and simply applying CNN may not achieve the best performance. Including regions with weak correlations to predict a target region actually hurts the performance. Thus, we use the local CNN to model the spatial dependency.

For each time interval $t$, we treat the target region $i$ and its surrounding neighbors as a $S\times S$ image with two channels $\mathbf{Y}_{i,t}\in \mathbb{R}^{S\times S\times 2}$. One channel contains start volume information, another one is end volume information. The target region is in the center of the image. The local CNN takes $\mathbf{Y}_{i,t}$ as input $\mathbf{Y}_{i,t}^{(0)}$, and the formulation of each convolutional layer $k$ is:
\begin{equation}
\label{eq:spatialcnn}
\mathbf{Y}_{i,t}^{(k)}={\rm ReLU}(\mathbf{W}^{(k)} \ast \mathbf{Y}_{i,t}^{(k-1)} +\mathbf{b}^{(k)}),
\end{equation}
where $\mathbf{W}^{(k)}$ and $\mathbf{b}^{(k)}$ are learned parameters. After stacking $K$ convolutional layers, a fully connected layer following a flatten layer is used to infer the spatial representation of region $i$ as $\mathbf{y}_{i, t}$.
\subsubsection{Short-term Temporal Dependency} 
We use Long Short-Term Memory (LSTM) network to capture the temporal sequential dependency, which is proposed to address the exploding and vanishing gradient issue of traditional Recurrent Neural Network (RNN). In this paper, we use the original version of LSTM~\cite{hochreiter1997long} and formulate it as:
\begin{equation}
\label{eq:lstm}
	\mathbf{h}_{i,t}=\mathrm{LSTM}([\mathbf{y}_{i,t};\mathbf{e}_{i,t}],\mathbf{h}_{i,t-1}),
\end{equation}
where $\mathbf{h}_{i,t}$ is the output representation of region $i$ at time interval $t$. $\mathbf{e}_{i,t}$ means external features (e.g., weather, event) and can be incorporated with $\mathbf{y}_{i,t}$ if applicable. Thus, the $\mathbf{h}_{i,t}$ contains both spatial and short-term temporal information. 

\subsection{Spatial Dynamic Similarity: Flow Gating Mechanism}
As we described before, local CNN is used to capture the spatial dependency. CNN handles the local structure similarity by local connection and weight sharing. In local CNN, the local spatial dependency relies on the similarity of historical traffic volume. However, the spatial dependency of volume is stationary, which can not fully reflect the relation between the target region and its neighbors. A more direct way to represent interactions between regions is traffic flow. If there are more flows existing between two regions, the relation between them is stronger (i.e., they are more similar). Traffic flow can be used to explicitly control the volume information propagation between regions. Therefore, we design a \textbf{F}low \textbf{G}ating \textbf{M}echanism (\textbf{FGM}), which explicitly capture dynamic spatial dependency in the hierarchy.
 
Similar to local CNN, we construct the local spatial flow image to protect the spatial dependency of flow. The traffic flow related to a certain region in a time interval falls into two categories, i.e., inflow departing from other location ending in the region during the time interval, and outflow starting from this region toward somewhere else. Two flow matrices for the region in this time interval can be constructed accordingly, where each element denotes inflow/outflow from/to other corresponding region. An example of outflow matrix is given in Figure \ref{fig:framework}(c).

Given a specific region $i$, we retrieve related traffic flow from past $l$ time intervals (i.e., time interval $t-l+1$ to $t$). The acquired flow matrices are further stacked and denoted by $\mathbf{F}_{i,t} \in  \mathbb{R}^{S\times S\times 2l}$, where $S\times S$ suggests the surrounding neighbor region size, and $2l$ is the number of flow matrices (two matrices for each time interval). Because the stacked flow matrices include all past flow interaction related to region $i$, 

we use CNN to model the spatial flow interactions between regions, which takes $\mathbf{F}_{i,t}$ as input $\mathbf{F}_{i,t}^{(0)}$. For each layer $k$, the formulation is 
\begin{equation}
\label{eq:flowconv}
\mathbf{F}_{i,t}^{(k)}=\mathrm{ReLU}(\mathbf{W}_{f}^{(k)} \ast \mathbf{F}_{i,t}^{(k-1)} +\mathbf{b}_{f}^{(k)}),
\end{equation}
where $\mathbf{W}_{f}^{(k)}$ and $\mathbf{b}_{f}^{(k)}$ are learned parameters.

At each layer, we use flow information to explicitly capture dynamic similarity between regions by constricting the spatial information via a flow gate. Specifically, the output of each layer is the spatial representation $\mathbf{Y}_t^{i,k}$ modulated by the flow gate. Formally, we revise the Eq.~\eqref{eq:spatialcnn} as: 
\begin{equation}
\mathbf{Y}_{i,t}^{(k)}={\rm ReLU}(\mathbf{W}^{(k)} \ast \mathbf{Y}_{i,t}^{(k-1)} +\mathbf{b}^{(k)})\otimes\sigma(\mathbf{F}_t^{i,k-1}),
\end{equation}
where $\otimes$ is the element-wise product between tensors. 

After $K$ gated convolutional layers, we use a flatten layer followed by a fully connected layer to get the flow gated spatial representation as $\mathbf{y}_{i,t}$. 

We replace the spatial representation $\mathbf{y}_{i,t}$ defined in Eq.~\eqref{eq:lstm} by $\mathbf{\hat{y}}_{i,t}$.

\subsection{Temporal Dynamic Similarity: Periodically Shifted Attention Mechanism}
In local spatial-temporal network defined above, only previous several time intervals (usually several hours) are used for prediction. However, it overlooks the long-term dependency (e.g., periodicity), which is an important property of spatial-temporal prediction problem~\cite{zonoozi2018periodic}. In this section, we take long-term periodic information into consideration.
\begin{figure}[t!]
	\centering
	\begin{subfigure}[b]{0.22\textwidth}
		\centering
		\includegraphics[height=0.7\textwidth]{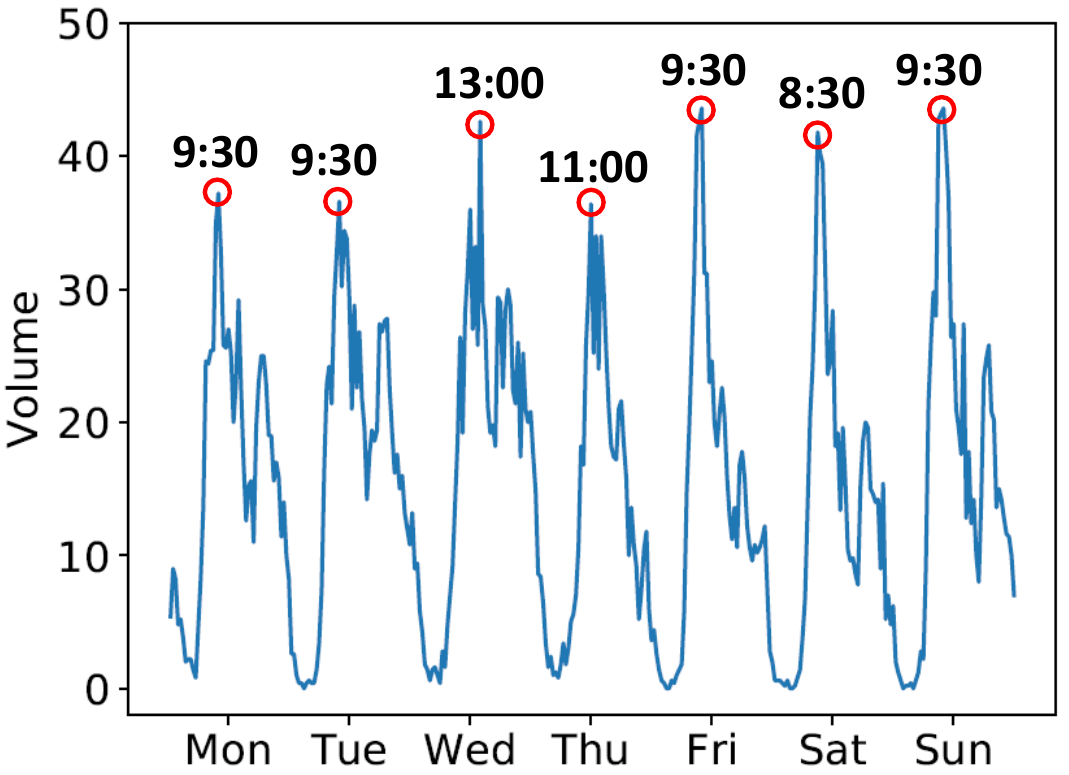}
		\caption{\label{fig:dayshift}}
	\end{subfigure}
	\begin{subfigure}[b]{0.22\textwidth}
		\centering
		\includegraphics[height=0.7\textwidth]{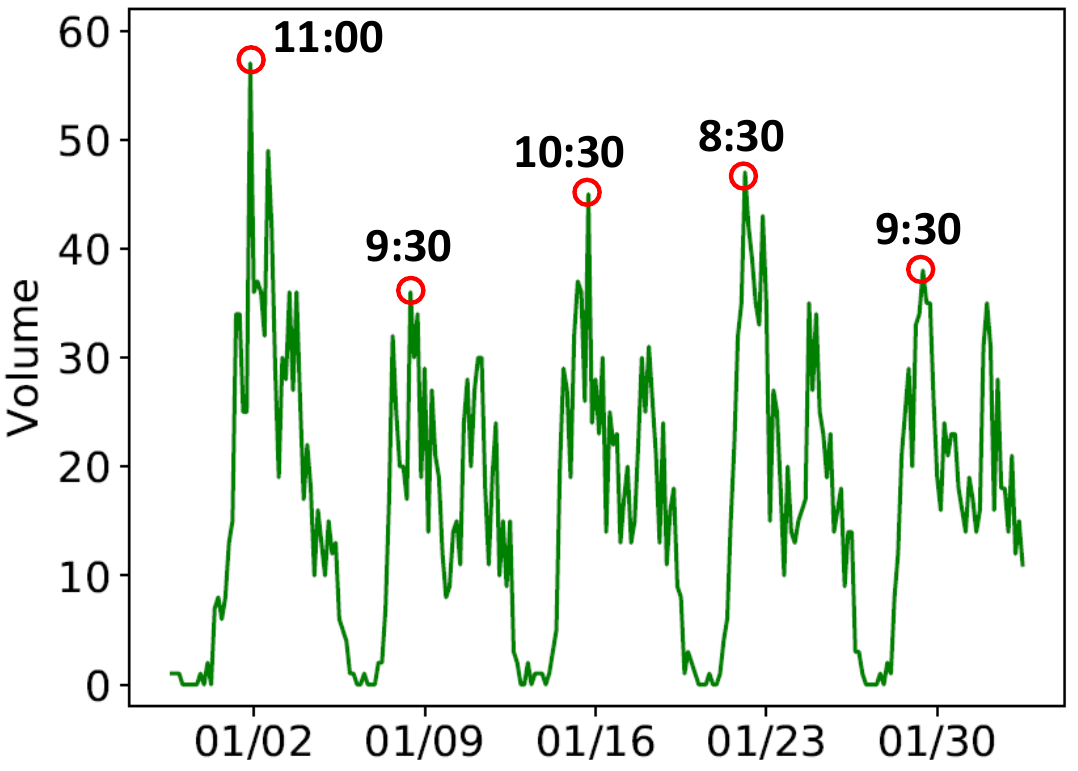}
		\caption{\label{fig:weekshift}}
	\end{subfigure}
	\vspace{-1em}
	\caption{The temporal shifting of periodicity. (a) Temporal shifting between different days. (b) Temporal shifting between different weeks. Note that, each time in these figures represents a time interval (e.g., 9:30am means 9:00-9:30am).}
\vspace{-1em}
\end{figure}

Training LSTM to handle long-term information is a nontrivial task, since the increasing length enlarges the risk of gradient vanishing, thus significantly weaken the effects of periodicity. To address this issue, relative time intervals of the predicting target (e.g., same time of yesterday, and the day before yesterday)  should be explicitly modeled. 
However, purely incorporating relative time intervals is insufficient ignores temporal shifting of periodicity, i.e., traffic data is not strictly periodic. For example, the peak hours on weekdays are usually in the afternoon, but could vary from 4:30pm to 6:00pm. Temporal shifting of periodic information is ubiquitous in traffic sequence because of accident or traffic congestions. An example of temporal shifting between different days and weeks is shown in Figure~\ref{fig:dayshift} and~\ref{fig:weekshift}, respectively. These two time series are start volume of the region containing Javits Center, calculated from New York Taxi Trips~\cite{nyctaxidata}. Clearly, the traffic series are periodic but the peaks of those series (i.e., marked by the red circle) exist in different time of the day. Besides, comparing these two figures, the periodicity is not strict daily or weekly. Thus, we design a \textbf{P}eriodically \textbf{S}hifted \textbf{A}ttention \textbf{M}echanism (\textbf{PSAM}) to tackle the limitations. The detailed approach is described as follows. 

We focus on addressing the shifting in daily periodicity. As shown in Figure~\ref{fig:framework}(a), relative time intervals from previous $P$ days are included for handling the periodic dependency. For each day, in order to tackle the temporal shifting problem, we further select $Q$ time intervals from each day in $Q$. For example, if the predicted time is 9:00-9:30pm, we select 1 hour before and after the predicted time (i.e., 8:00-10:30pm and $|Q|=5$). These time intervals $q \in Q$ are used to tackle the potential temporal shifting. Additionally, we use LSTM to protect the sequential information for each day $p \in P$, which is formulated as:

\begin{equation}
	\mathbf{h}_{i,t}^{p,q}=\mathrm{LSTM}([\mathbf{y}_{i,t}^{p,q};\mathbf{e}_{i,t}^{p,q}],\mathbf{h}_{i,t}^{p,q-1}),
\end{equation}
where $\mathbf{h}_{i,t}^{p,q}$ is the representation of time $q$ in previous day $p$ for the predicted time $t$ in region $i$.

We adopt an attention mechanism to capture the temporal shifting and get the weighted representation of each previous day. Formally, the representation of each previous days $\mathbf{h}_{i,t}^{p}$ is a weighted sum of representations in each selected time interval $q$, which is defined as:
\begin{equation}
\mathbf{h}_{i,t}^{p}=\sum_{q \in Q} \alpha_{i,t}^{p,q} \mathbf{h}_{i,t}^{p,q},
\end{equation}
where weight $\alpha_{i,t}^{p,q}$ measures the importance of the time interval $q$ in day $p \in P$. The importance value $\alpha_{i,t}^{p,q}$ is derived by comparing the learned spatial-temporal representation from short-term memory (i.e., Eq.~\eqref{eq:lstm}) with previous hidden state $\mathbf{h}_{i,t}^{p,q}$. Formally, the weight $\alpha_{i,t}^{p,q}$ is defined as
\begin{equation}
\alpha_{i,t}^{p,q}=\frac{\exp(\rm{score}(\mathbf{h}_{i,t}^{p,q},\mathbf{h}_{i,t}))}{\sum_{q \in Q} \exp(\rm{score}(\mathbf{h}_{i,t}^{p,q},\mathbf{h}_{i,t}))}.
\end{equation}
In this work, similar to~\cite{luong2015effective}, the score function is regarded as content-based function defined as:
\begin{equation}
\rm{score}(\mathbf{h}_{i,t}^{p,q},\mathbf{h}_{i,t})=\mathbf{v}^T\tanh(\mathbf{W_H}\mathbf{h}_{i,t}^{p,q}+\mathbf{W_X}\mathbf{h}_{i,t}+\mathbf{b_X}),
\end{equation}
where $\mathbf{W_H}$, $\mathbf{W_X}$, $\mathbf{b_X}$, $\mathbf{v}$ are learned parameters, $\mathbf{v}^T$ denotes the transpose of $\mathbf{v}$. For each previous day $p$, we get the periodic representation $\mathbf{h}_{i,t}^{p}$. Then, we use another LSTM to preserve the sequential information by using these periodic representations as input, i.e.,
\begin{equation}
	\hat{\mathbf{h}}_{i,t}^{p}=\mathrm{LSTM}(\mathbf{h}_{i,t}^{p}, \hat{\mathbf{h}}_{i,t}^{p-1}).
\end{equation}
We regard the output of the last time interval $\hat{\mathbf{h}}_{i,t}^{P}$ as the representation of temporal dynamic similarity (i.e., long-term periodic information).

\subsection{Joint Training}
We concatenate the short-term representation $\mathbf{h}_{i,t}$ and long-term representation $\hat{\mathbf{h}}_{i,t}^{P}$ as $\mathbf{h}_{i,t}^{c}$, which preserve both short-term and long-term dependencies for predicting region and time.
Then we feed $\mathbf{h}_{i,t}^{c}$ to a fully connected layer and get the final prediction value of start and end traffic volume for each region $i$, which is denoted as $y^i_{s,t+1}$ and $y^i_{e,t+1}$, respectively. The final prediction function is defined as: 
\begin{equation}
[y_{i,t+1}^{s}, y_{i,t+1}^{e}]=\tanh(\mathbf{W}_{fa}\mathbf{h}_{i,t}^{c}+\mathbf{b}_{fa}),
\end{equation}
where $\mathbf{W}_{fa}$ and $\mathbf{b}_{fa}$ are learned parameters. The output of our model is (-1,1) since we normalize the value of start and end volume. We later denormalize the prediction to get the actual demand values.

In this work, we predict start volume and end traffic volume simultaneously, the loss function is defined as:
\begin{equation}
\label{eq:loss}
\mathcal{L} = \sum_{i=1}^n \lambda(y_{i,t+1}^{s} - \hat{y}_{i,t+1}^{s})^2+(1-\lambda)(y_{i,t+1}^{e} - \hat{y}_{i,t+1}^{e})^2,
\end{equation}
where $\lambda$ is a parameter to balance the influence of start and end. The actual value of start and end volume in region $i$ at time $t+1$ are denoted as: $\hat{y}_{i,t+1}^{s}$, $\hat{y}_{i,t+1}^{e}$.

\section{Experiment}
\label{sec:experiment}
\subsection{Experiment Settings}
\subsubsection{Datasets}
We evaluate our proposed method on two large-scale public real-world datasets from New York City (NYC). Each dataset contains trip records, as detailed follows.
\begin{itemize}[leftmargin=*]
	\item \textbf{NYC-Taxi:} NYC-Taxi dataset contains $22,349,490$ taxi trip records of NYC~\cite{nyctaxidata} in 2015, from 01/01/2015 to 03/01/2015. In the experiment, we use data from 01/01/2015 to 02/10/2015 (40 days) as training data, and the remained 20 days as testing data.
	\item \textbf{NYC-Bike:} The bike trajectories are collected from NYC Citi Bike system~\cite{nycbikedata} in 2016, from 07/01/2016 to 08/29/2016. The dataset contains $2,605,648$ trip records. The previous 40 days (i.e., from 07/01/2016 to 08/09/2016) are used as training data, and the rest 20 days as testing data. 
\end{itemize}
\subsubsection{Preprocessing}
We split the whole city as 10$\times$20 regions. The size of each region is about $1km\times 1km$. The length of each time interval is set as 30 minutes. We use Min-Max normalization to convert traffic volume and flow to $[0,1]$ scale. After prediction, we denormalize the prediction value and use it for evaluation. We use a sliding window on both training and testing data for sample generation. When testing our model, we filter the samples with volume values less than 10, which a common practice used in industry and academy~\cite{yao2018deep}. Because in the real-world applications, cares with low traffic are of little interest. We select 80\% of the training data to learn the models, and the remaining 20\% for validation. 

\subsubsection{Evaluation Metric \& Baselines}
\label{sec:baseline}
In our experiment, two commonly metrics are used for evaluation: (1) Mean Average Percentage Error (MAPE) (2) Rooted Mean Square Error (RMSE). We compare STDN with widely used time series regression models, including (1) Historical average (HA) (2) Autoregressive integrated moving average (ARIMA); The following traditional regression methods are included: (3) Ridge Regression (Ridge); (4) LinUOTD~\cite{tong2017sim}; (5) XGBoost~\cite{chen2016xgboost}. In addition, neural-network-based methods are also considered: (6) MultiLayer Perceptron (MLP); (7) Convolutional LSTM (ConvLSTM)~\cite{xingjian2015convolutional}; (8) DeepSD~\cite{wang2017deepsd}; (9) Deep Spatio-Temporal Residual Networks (ST-ResNet)~\cite{zhang2016deep}; (10) Deep Multi-View Spatial-Temporal Network (DMVST-Net)~\cite{yao2018deep}. 
\subsubsection{Hyperparameter Settings}
\label{sec:parasetting}
We set the hyperparameters based on the performance on validation set. For spatial information, we set all convolution kernel sizes to $3\times 3$ with 64 filters. The size of each neighborhood considered was set as $7\times7$. We set $K=3$ (number of layers), $l=2$ (the time span of considered flow). For temporal information, we set the length of short-term LSTM as 7 (i.e., previous 3.5 hours), $|P|=3$ for long-term periodic information (i.e., previous 3 days), $|Q|=3$ for periodically shifted attention mechanism (i.e., half an hour before and after of relative predicted time are considered), the dimension of hidden representation of LSTM is 128. STDN is optimized via and Adam~\cite{kingma2014adam}. The batch size in our experiment is set to 64. Learning rate is set as 0.001. Both dropout and recurrent dropout rate in LSTM are set as 0.5. We also use early-stop in all the experiments. $\lambda$ is set as $0.5$ to balance start and end volume. 

\subsection{Results}
\begin{table*}[t!]
	\centering
	\caption{Comparison with Different Baselines}
	\label{tab:baseline_taxi}
	\begin{tabular}{l||l|c|c|c|c}
		\hline
		\multicolumn{1}{l||}{\multirow{2}{*}{Dataset}} & \multicolumn{1}{l|}{\multirow{2}{*}{Method}} & \multicolumn{2}{c}{Start} & \multicolumn{2}{|c}{End} \\\cline{3-6}
		&                  & RMSE      & MAPE      & RMSE      & MAPE      \\\hline\hline
		\multicolumn{1}{l||}{\multirow{11}{*}{NYC-Taxi}} & HA                          &    43.82         &     23.18\%        &         33.83     &    21.14\%           \\
		& ARIMA                                       &     36.53         &       22.21\%        &      27.25        &         20.91\%      \\
		& LR                          &        28.51      &        19.94\%       &       24.38       &         20.07\%      \\
		&MLP                   &     26.67$\pm$0.56         &      18.43$\pm$0.62\%         &        22.08$\pm$0.50      &      18.31$\pm$0.83\%         \\
		&XGBoost       &     26.07         &      19.35\%        &          21.72    &       18.70\%        \\
		&LinUOTD  & 28.48 & 19.91\% & 24.39 &20.03\%\\
		&ConvLSTM  & 28.13$\pm$0.25 & 20.50$\pm$0.10\% & 23.67$\pm$0.20 &  20.70$\pm$0.20\%\\
		&DeepSD  & 26.35$\pm$0.53 & 18.12$\pm$0.38\% & 21.95$\pm$0.35 & 18.15$\pm$0.62\%\\
		&ST-ResNet       &       26.23$\pm$0.33       &    21.13$\pm$0.63\%           &       21.63$\pm$0.25       &  21.09$\pm$0.51\%             \\
		&DMVST-Net         &       25.74$\pm$0.26       &    17.38$\pm$0.46\%           &        20.51$\pm$0.46      &        17.14$\pm$0.32\%       \\\cline{2-6}
		&STDN                                       &      \textbf{24.10$\pm$0.25}***        &         \textbf{16.30$\pm$0.23\%}***      &      \textbf{19.05$\pm$0.31}***        &   \textbf{16.25$\pm$0.26\%}***  \\\hline \hline 
		\multicolumn{1}{l||}{\multirow{11}{*}{NYC-Bike}} & HA                          &    12.49         &     27.82\%          &         11.93     &    27.06\%           \\
		& ARIMA                                       &         11.53     &      26.35\%         &        11.25      &        25.79\%               \\
		& LR                          &        10.92      &       25.29\%        &        10.33      &        24.58\%       \\
		& MLP                   &       9.83$\pm$0.19       &        23.12$\pm$0.47\%       &     9.12$\pm$0.24         &         22.40$\pm$0.40\%      \\
		& XGBoost       &        9.57      &     23.52\%          &        8.94      &        22.54\%       \\
		& LinUOTD & 11.04 & 25.22\% & 10.44 & 24.44\%\\
		& ConvLSTM  & 10.40$\pm$0.17 & 25.10$\pm$0.45\% & 9.22$\pm$0.19 & 23.20$\pm$0.47\% \\
		& DeepSD & 9.69 & 23.62\% & 9.08 & 22.36\%\\
		& ST-ResNet       &      9.80$\pm$0.12      &       25.06$\pm$0.36\%        &       8.85$\pm$0.13       &      22.98$\pm$0.53\%         \\
		& DMVST-Net         &     9.14$\pm$0.13       &      22.20$\pm$0.33\%           &     8.50$\pm$0.19         &         21.56$\pm$0.49\%      \\\cline{2-6}
		& STDN                                       &      \textbf{8.85$\pm$0.11}***        &         \textbf{21.84$\pm$0.36\%}**      &      \textbf{8.15$\pm$0.15}***        &   \textbf{20.87$\pm$0.39\%}***  \\\hline
	\end{tabular}
	\\\vspace{0.1cm}
	 *** (**) means the result is significant according to Student’s T-test at level 0.01 (0.05) compared to DMVST-Net
\end{table*}
\subsubsection{Performance Comparison}
\label{sec:compare_baseline}
Table~\ref{tab:baseline_taxi} show the performance of our proposed method as compared to all other competing methods in NYC-Taxi and NYC-Bike datasets, respectively. We run each baseline 10 times and report the mean and standard deviation of each baseline. Besides, we also conduct student t-test. Our proposed STDN significantly outperforms all competing baselines by achieving the lowest RMSE and MAPE on both datasets.

Specifically, the traditional time-series prediction methods (HA and ARIMA) do not perform well, because they only rely on historical records of predicting value and overlook spatial and other context features. For regression-based methods (Ridge, LinUOTD, XGBoost), they further consider spatial correlations as features or regularizations. As a result, they achieve better performances compared with other conventional time-series approaches. However, they fail to capture the complex non-linear temporal dependencies and the dynamic spatial relationships. Therefore, our proposed method significantly outperforms those regression-based methods.

For neural-network-based methods, STDN outperforms MLP and DeepSD. The potential reason is that MLP and DeepSD do not explicitly model spatial dependency and temporal sequential dependency. Also, our model outperforms ST-ResNet, because ST-ResNet uses CNN to capture spatial information, but overlooks the temporal sequential dependency. ConvLSTM extends fully connected LSTM by integrating convolutional operation to LSTM units for capturing both spatial and temporal information. DMVST-Net considers spatial-temporal information by local CNN and LSTM. However, these two models overlook the dynamic spatial similarity and periodic temporal shifting. The better performance of our proposed model demonstrates the effectiveness of flow gating mechanism and periodically shifted attention mechanism to capture the dynamic spatial-temporal similarity. 
\subsubsection{Effectiveness of Flow Gating Mechanism}
\label{sec:flowgateexperiment}
In this section, we study the effectiveness of flow gating mechanism. We first list some variants of using traffic flow information as follows:
\begin{itemize}[leftmargin=*]
	\item \textbf{LSTN}: As described in Section~\ref{sec:localspatial}, only short-term temporal dependency, and local spatial dependency are considered.
	\item \textbf{LSTN-FI}: LSTN-FI use traffic flow information as features. We simply concatenate flow information $\mathbf{F}_{i,t}$ defined in Eq.~\eqref{eq:flowconv} and spatial representation $\mathbf{Y}_{i,t}$. Then we feed it in to LSTM as spatial features instead of using a flow gating mechanism.
	\item \textbf{LSTN-FGM}: FGLSTN further utilize flow gating mechanism to represent the spatial dynamic similarity between local neighborhoods. The variant does not use periodically shifted attention mechanism.
\end{itemize}

\begin{figure}[!t]
	\centering
	\begin{subfigure}[b]{0.22\textwidth}
		\centering
		\includegraphics[height=0.8\textwidth]{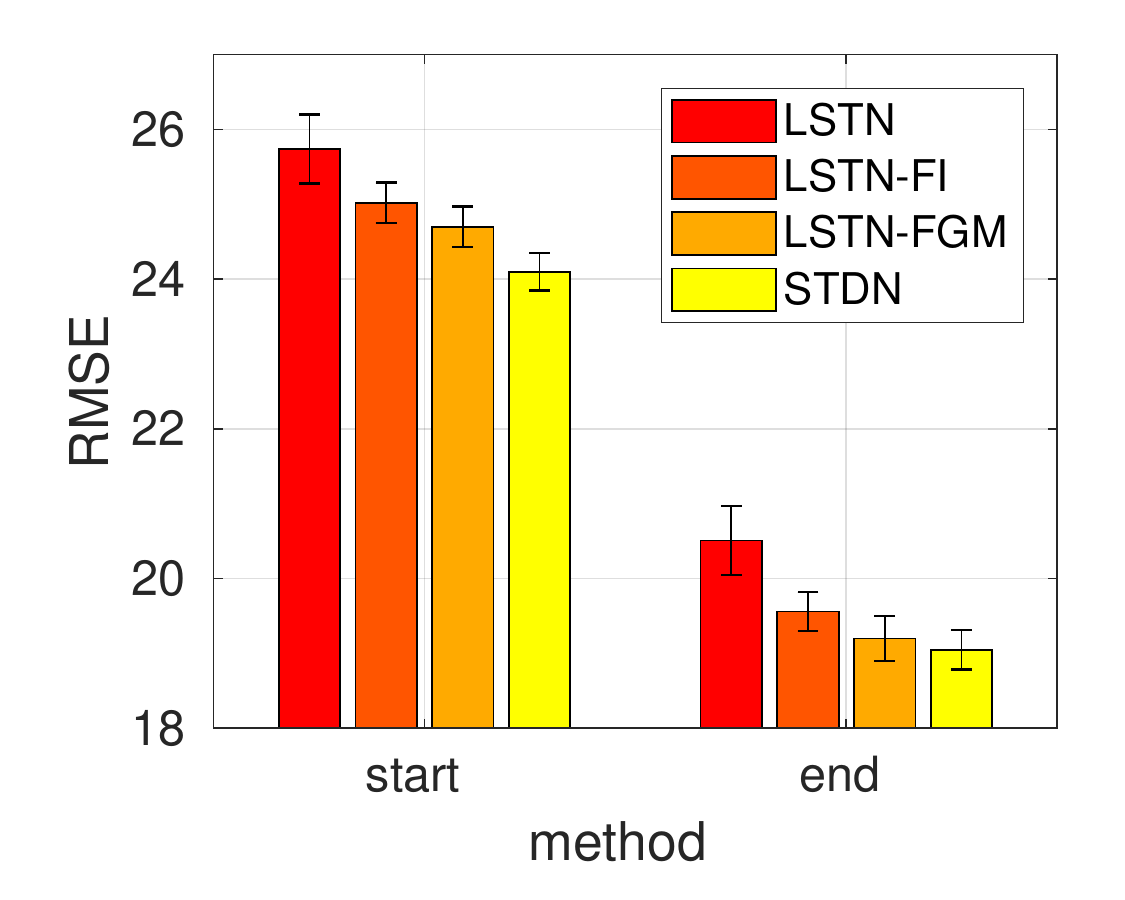}
		\caption{\label{fig:taxiflowrmse}: RMSE on NYC-Taxi}
	\end{subfigure}
	\begin{subfigure}[b]{0.22\textwidth}
		\centering
		\includegraphics[height=0.8\textwidth]{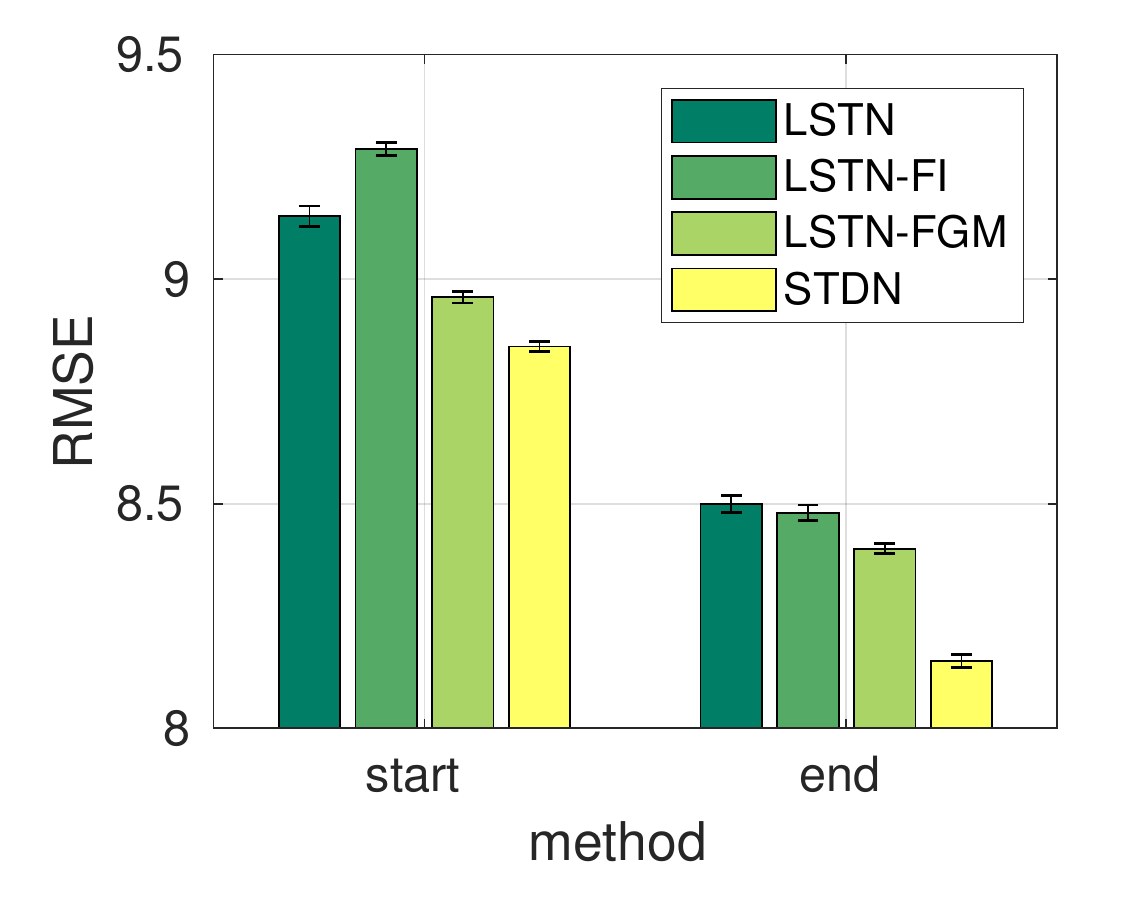}
		\caption{\label{fig:bikeflowrmse}: RMSE on NYC-Bike}
	\end{subfigure}
    \begin{subfigure}[b]{0.22\textwidth}
		\centering
		\includegraphics[height=0.8\textwidth]{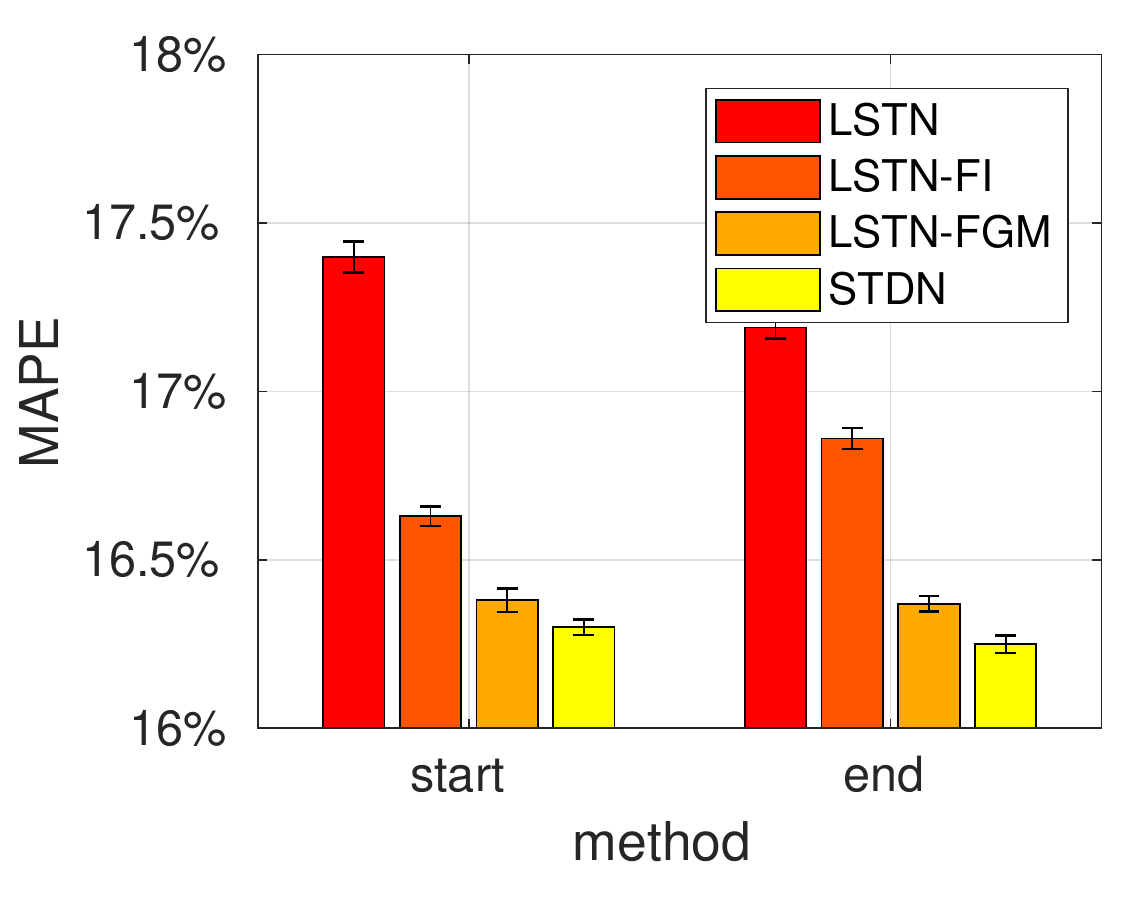}
		\caption{\label{fig:taxiflowmape}: MAPE on NYC-Taxi}
	\end{subfigure}
	\begin{subfigure}[b]{0.22\textwidth}
		\centering
		\includegraphics[height=0.8\textwidth]{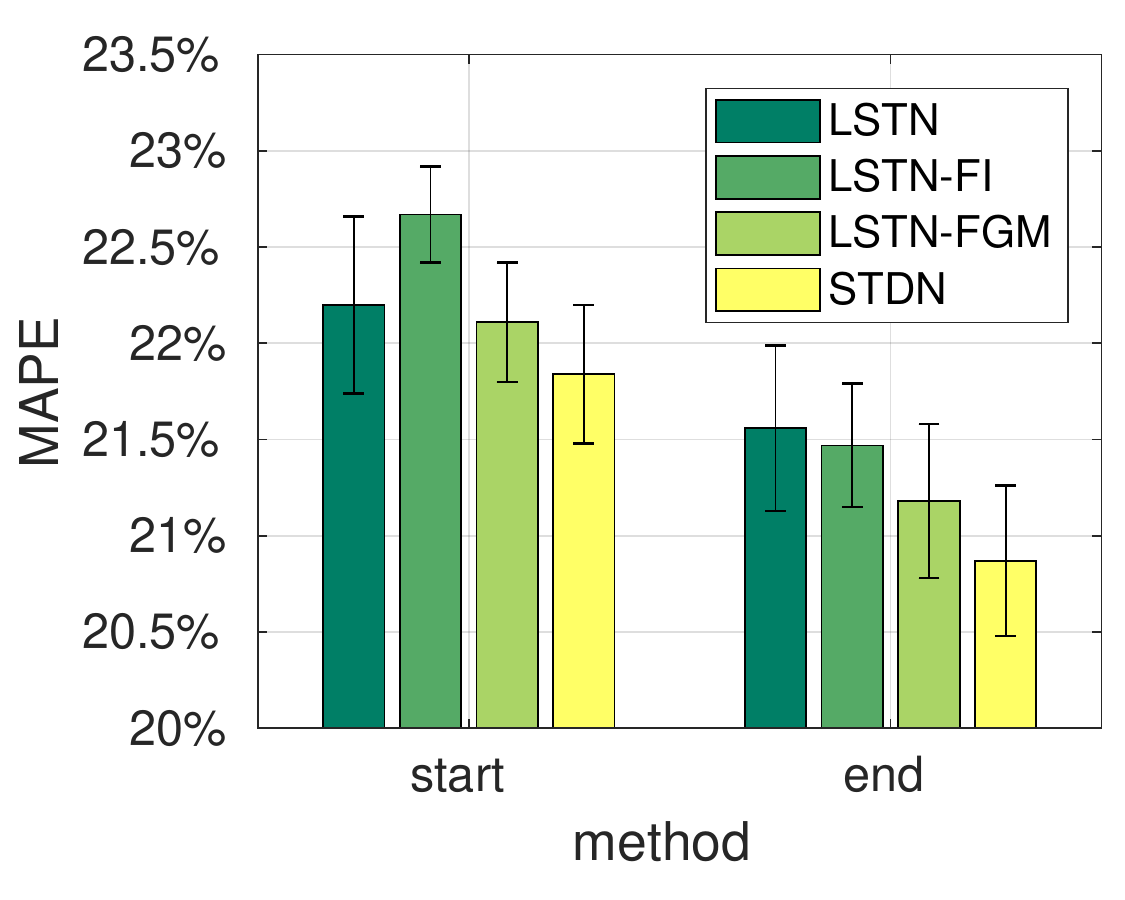}
		\caption{\label{fig:bikeflowmape}: MAPE on NYC-Bike}
	\end{subfigure}
	\caption{Evaluation of flow gating mechanism (FGM) and its variants.} 
\end{figure}
The results of different variants in NYC-Taxi an NYC-Bike are shown in Figure~\ref{fig:taxiflowrmse},~\ref{fig:taxiflowmape} and Figure~\ref{fig:bikeflowrmse},~\ref{fig:bikeflowmape}, respectively. LSTN-FGM and STDN outperform LSTN, because LSTN overlooks the dynamic spatial similarity between regions (e.g., traffic flow). In order to model dynamic spatial similarity, a straightforward approach would be using local flow as another type of spatial representation, i.e., the variants LSTN-FI. However, compared to LSTN-FGM, which uses flow gating mechanism, LSTN-FI performs worse. One potential reason is that only using traffic flow as features can not incorporate the structure of spatial dynamic similarity.
The results reveal the effectiveness of flow gating mechanism to explicitly capture the dynamic spatial similarity. Furthermore, the comparison to STDN demonstrates the importance of tackle temporal shifted periodic information.

\subsubsection{Effectiveness of Periodically Shifted Attention Mechanism}
The intuition of periodically shifted attention mechanism is the long-term periodic information and temporal shifting. In this section, we analyze the effectiveness of periodically shifted attention mechanism and several variants are listed as follows:
\begin{itemize}[leftmargin=*]
		\item \textbf{LSTN-L}: We extend LSTN by taking long-term sequential information into consideration. The long-term information (i.e., the information of relative predicted time in previous 3 days) are concatenated with short-term information (i.e., the information of previous 7 time intervals) and use one LSTM network as prediction component.
		\item \textbf{LSTN-SL}: LSTN-SL removes the periodically shifted attention mechanism in STDN. LSTN-SL consists of two LSTM network. One is used to capture short-term dependency, and another one uses relative time in previous 3 days information to capture long-term information. Note that we set $|Q|=1$ (only relative predicted time in previous 3 days are considered) and LSTN-SL does not include flow gating mechanism.
		\item \textbf{LSTN-PSAM}: We add the periodically shifted attention mechanism attention on LSTN-SL. Compared to proposed STDN, this variant only removes the flow gating mechanism.
\end{itemize}

\begin{figure}[!t]
	\vspace{-1em}
	\centering
	\begin{subfigure}[b]{0.22\textwidth}
		\centering
		\includegraphics[height=0.8\textwidth]{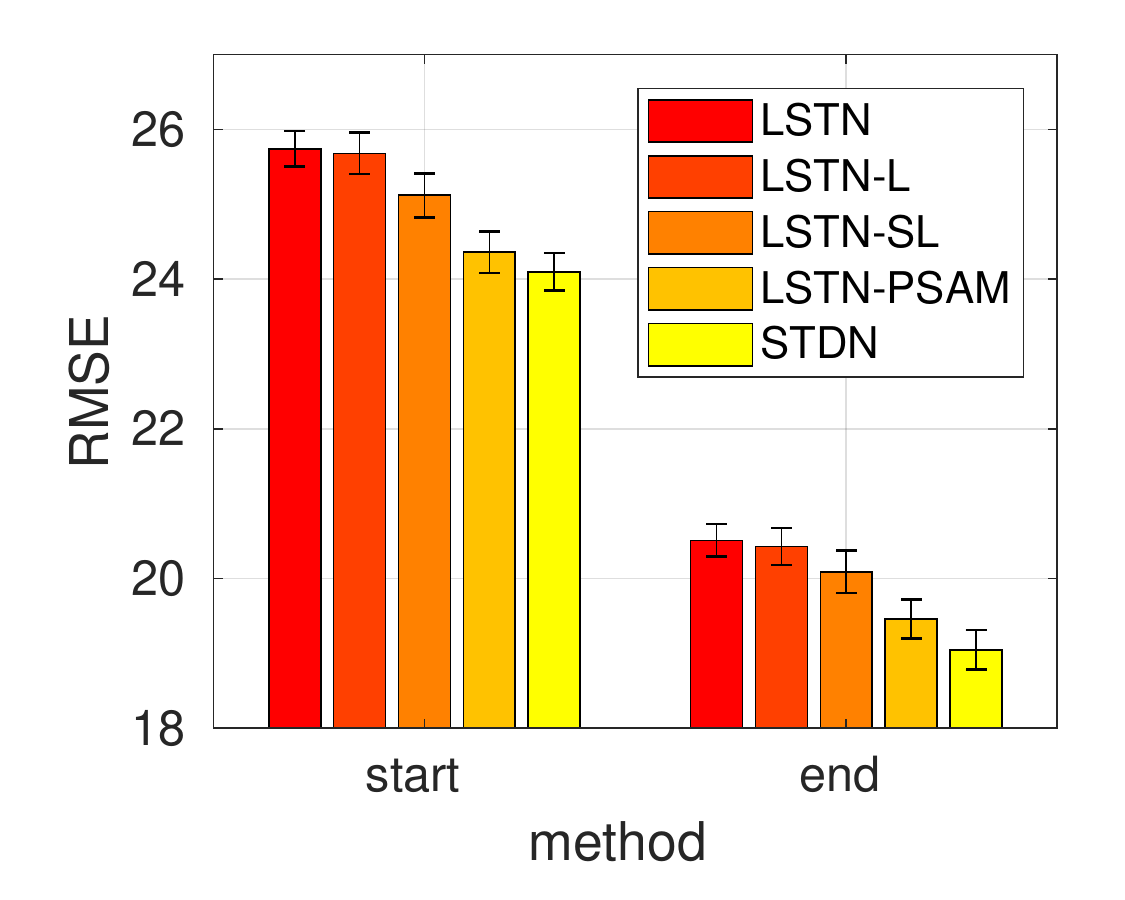}
		\caption{\label{fig:taxitimermse}: RMSE on NYC-Taxi}
	\end{subfigure}
	\begin{subfigure}[b]{0.22\textwidth}
		\centering
		\includegraphics[height=0.8\textwidth]{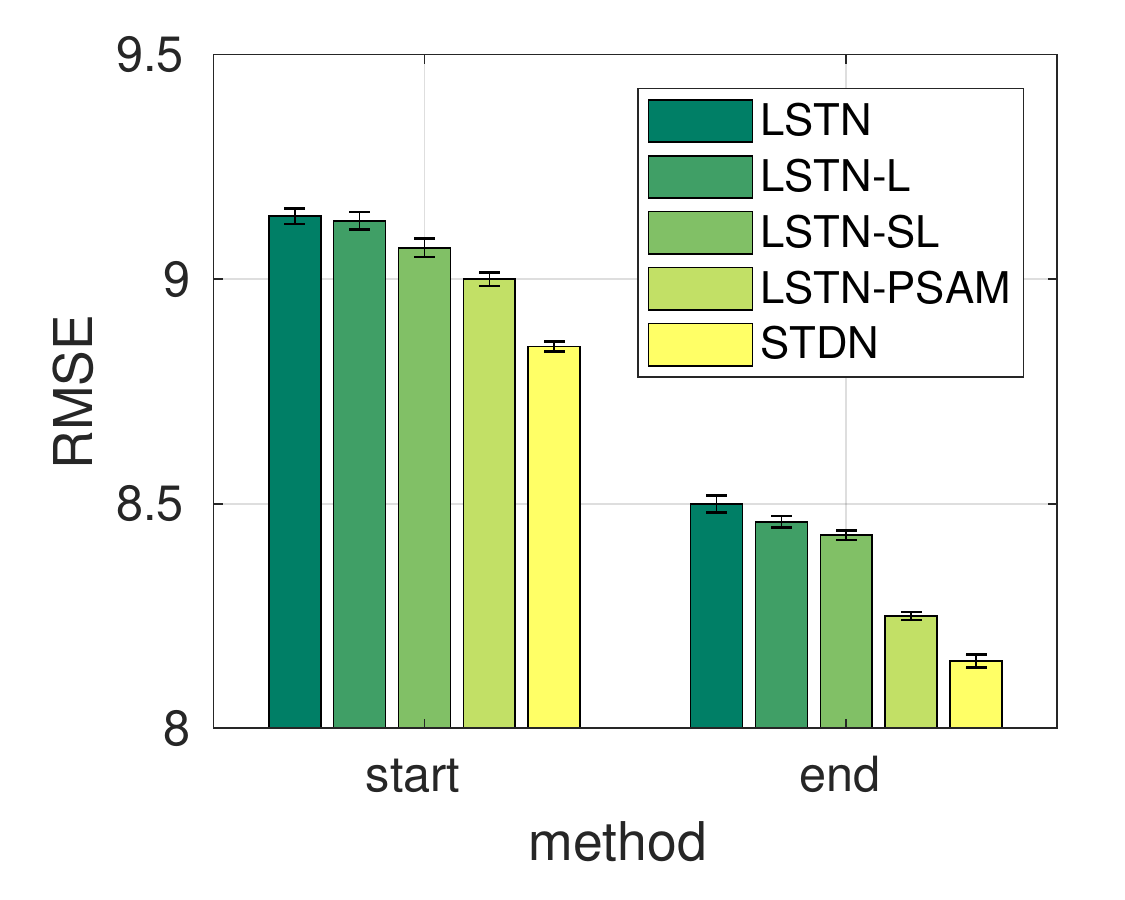}
		\caption{\label{fig:biketimermse}: RMSE on NYC-Bike}
	\end{subfigure}
    \begin{subfigure}[b]{0.22\textwidth}
		\centering
		\includegraphics[height=0.8\textwidth]{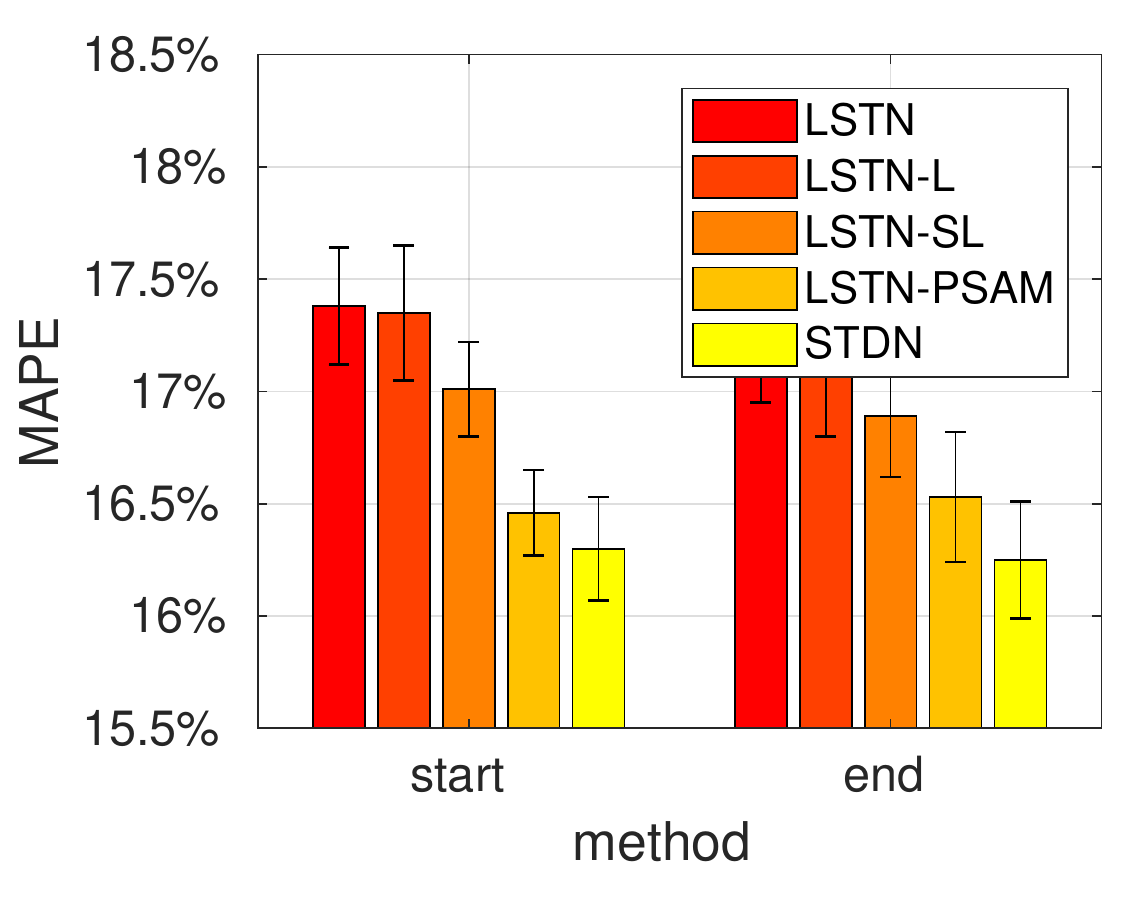}
		\caption{\label{fig:taxitimemape}: MAPE on NYC-Taxi}
	\end{subfigure}
	\begin{subfigure}[b]{0.22\textwidth}
		\centering
		\includegraphics[height=0.8\textwidth]{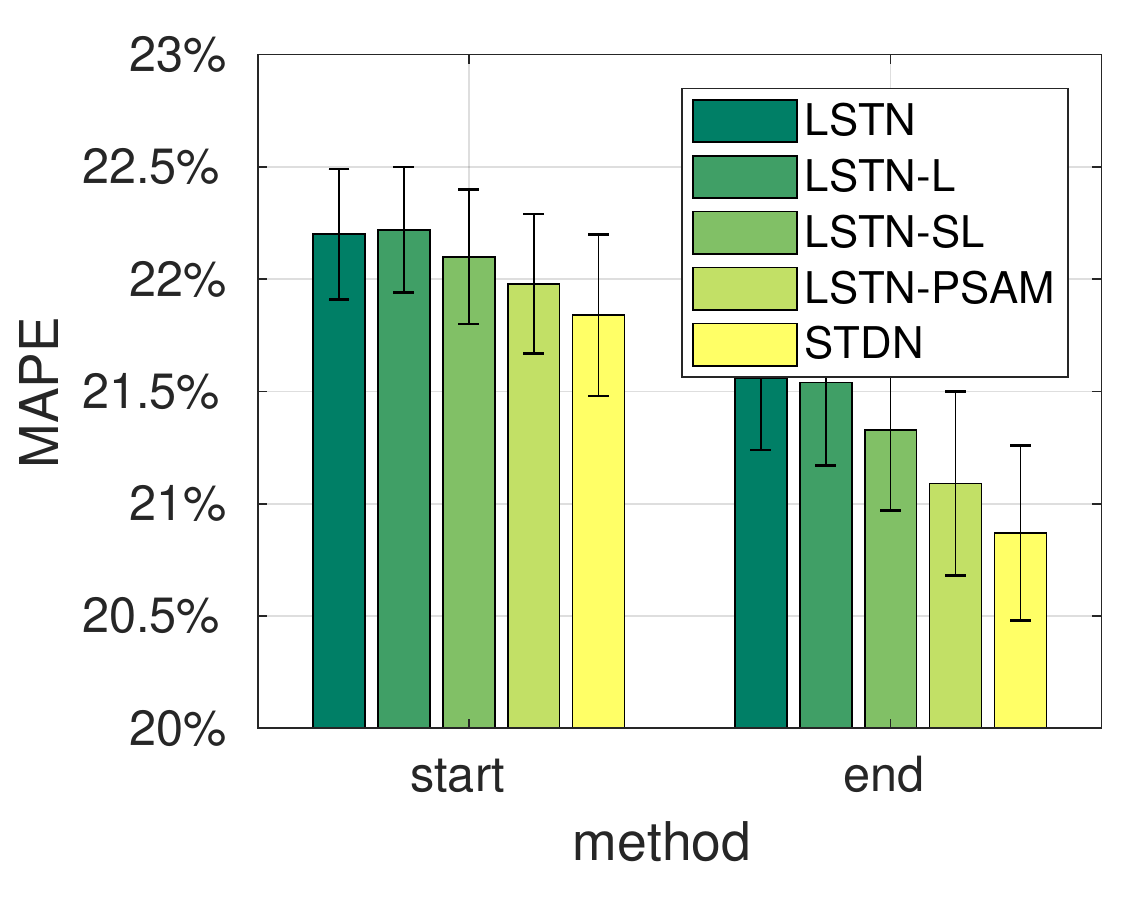}
		\caption{\label{fig:biketimemape}: MAPE on NYC-Bike}
	\end{subfigure}
	\caption{Evaluation of periodically shifted attention mechanism (PSAM) and its variants.}
    \vspace{-1em}
\end{figure}

Figures~\ref{fig:taxitimermse},~\ref{fig:taxitimemape} and Figures~\ref{fig:biketimermse},~\ref{fig:biketimemape} show the comparison results in NYC-Taxi and NYC-Bike, respectively. We also show LSTN and STDN (our proposed model) for comparison. The results for LSTN and LSTN-L are similar. 
One potential reason is that when the long term information is concatenated before short term information in LSTM, only the short term information can be remembered.
The other reason is the uneven time gap between long term information and short term information might be harmful for learning the periodic sequence. In one LSTM network, sequences with different sample rate may not achieve good performance. LSTN-SL further split the long-term and short-term information and use two LSTM networks to handle these dependencies. We can see that LSTN-SL performs better than LSTN-L, which shows the effectiveness of considering long-term and short-term information separately. Furthermore, the improvement from LSTN-PSAM to LSTN-SL shows the influence of temporal shifting. Using the proposed periodically shifted attention mechanism can capture the temporal shifting and improve the performance. Finally, the better performance of STDN than LSTN-PSAM further shows the effectiveness of flow gating mechanism. 

\section{Conclusion and Discussion}
\label{sec:conclusion}
In this paper, we propose a novel Spatial-Temporal Dynamic Network (STDN) for traffic prediction. Our approach tracks the dynamic spatial similarity between regions by flow gating mechanism and temporal periodic similarity by periodically shifted attention mechanism. The evaluation on two large-scale datasets show that proposed model outperforms the state-of-the-art methods. In the future, we plan to investigate the proposed model on other spatial-temporal prediction problems. In addition, we plan to explain the model (i.e., explain feature importance of traffic prediction), which is important for policy makers. Data and code can be found in https://github.com/tangxianfeng/STDN

\section{Acknowledgments}
The work was supported in part by NSF awards \#1544455, \#1652525,
\#1618448, and \#1639150. The views and conclusions contained in
this paper are those of the authors and should not be interpreted
as representing any funding agencies.
\bibliographystyle{aaai}
\fontsize{9.0pt}{10.0pt} \selectfont
\bibliography{ref} 
\end{document}